\documentclass[preprint,12pt]{elsarticle}
\usepackage{amssymb}
\journal{SoftwareX}

\usepackage{adjustbox}
\usepackage{caption,subcaption}

\usepackage{hyperref}
\hypersetup{breaklinks=true}
\usepackage{url}
\usepackage{breakurl}
\usepackage{lastpage}
\usepackage{amsmath}
\usepackage{xcolor}
\usepackage{listings}
\usepackage[simplified]{pgf-umlcd}
\usepackage{tikz}
\usetikzlibrary{positioning,calc,shapes.geometric}
\tikzset{
    every node/.style={font=\footnotesize},
    between/.style args={#1 and #2}{
         at = ($(#1)!0.5!(#2)$)
    }
}
\definecolor{codegreen}{rgb}{0,0.6,0}
\definecolor{codegray}{rgb}{0.5,0.5,0.5}
\definecolor{codepurple}{rgb}{0.58,0,0.82}
\definecolor{backcolour}{rgb}{0.98,0.98,0.96}
\lstdefinestyle{mystyle}{
    backgroundcolor=\color{backcolour},   
    commentstyle=\color{codegreen},
    keywordstyle=\color{magenta},
    numberstyle=\tiny\color{codegray},
    stringstyle=\color{codepurple},
    basicstyle=\ttfamily\scriptsize,
    breakatwhitespace=false,         
    breaklines=true,                 
    captionpos=b,                    
    keepspaces=true,                 
    numbers=none,                    
    numbersep=5pt,                  
    showspaces=false,                
    showstringspaces=false,
    showtabs=false,                  
    tabsize=2
}
\usepackage{pifont}
\usepackage{bm}
\definecolor{darkred}{RGB}{204,0,0}
\newcommand{\ck}{\textcolor{blue}{\ding{52}}}
\newcommand{\ckp}{\textcolor{blue}{\bf (\ding{52})}}
\newcommand{\ckm}{\textcolor{darkred}{$\bm{\exists$}}}
\newcommand{\xm}{\textcolor{darkred}{\ding{55}}}
\newcommand{\python}{\textcolor{blue}{\bf Python}}
\newcommand{\java}{\textcolor{darkred}{\bf Java}}
\newcommand{\csharp}{\textcolor{darkred}{\bf C\#}}
\newcommand{\bl}[1]{\textcolor{blue}{\bf #1}}
\newcommand{\rd}[1]{\textcolor{darkred}{\bf #1}}

\usepackage{bold-extra}
\newcommand{\tb}[1]{\texttt{\textbf{#1}}}

\begin{document}

\begin{frontmatter}

\title{EC-KitY: Evolutionary Computation Tool Kit in Python with Seamless Machine Learning Integration}

\author[cs]{Moshe Sipper}
\author[cs]{Tomer Halperin}
\author[cs]{Itai Tzruia}
\author[sise]{Achiya Elyasaf}

\affiliation[cs]{
    organization={Department of Computer Science, Ben-Gurion University of the Negev},
    city={Beer-Sheva},
    postcode={8410501}, 
    country={Israel}}

\affiliation[sise]{
    organization={Department of Software and Information Systems Engineering, Ben-Gurion University of the Negev},
    city={Beer-Sheva},
    postcode={8410501}, 
    country={Israel}}

\begin{abstract}
\tb{EC-KitY} is a comprehensive Python library for doing evolutionary computation (EC), licensed under the BSD 3-Clause License, and compatible with \tb{scikit-learn}. Designed with modern software engineering and machine learning integration in mind, \tb{EC-KitY} can support all popular EC paradigms, including genetic algorithms, genetic programming, coevolution, evolutionary multi-objective optimization, and more. This paper provides an overview of the package, including the ease of setting up an EC experiment, the architecture, the main features, and a comparison with other libraries. 
\end{abstract}

\begin{keyword}
Evolutionary Algorithms \sep Evolutionary Computation \sep Genetic Programming \sep Machine Learning \sep scikit-learn

\end{keyword}

\end{frontmatter}

\section{Introduction}
In Evolutionary Computation (EC)---or Evolutionary Algorithms (EAs)---core concepts from evolutionary biology—inheritance, random variation, and selection—are harnessed in algorithms that are applied to complex computational problems. As discussed by \citet{Sipper2017ec}, EAs present several important benefits over popular machine learning (ML) methods, including: less reliance on the existence of a known or discoverable gradient within the search space; ability to handle design problems, where the objective is to design new entities from scratch; fewer required a priori assumptions about the problem at hand; seamless integration of human expert knowledge; ability to solve problems where human expertise is very limited; support of interpretable solution representations; support of multiple objectives. 

Importantly, these strengths often dovetail with weak points of ML algorithms, which has resulted in an increasing number of works that fruitfully combine the fields of EC and ML or deep learning (DL). For example, 
\citet{Sipper2021cml} ``converted'' a selection method in EC to a random forest ensemble producer;
\citet{Lapid2022} used EC to evolve activation functions for DL-based image classifiers;
\citet{Lapid2022adv} designed an evolutionary algorithm for generating adversarial instances in deep convolutional neural networks; 
\citet{livne2022evolving} utilized EC and DL to create accurate and interpretable context-aware recommender systems. 
Major conferences in the field of EC now regularly address the combining of EC and ML, e.g., GECCO, arguably the major EC event, devotes two entire tracks and three workshops to the blending of evolution and learning. There are a number of good surveys on evolutionary machine learning: \citet{Telikani2022}, \citet{AlSahaf2019}, and \citet{Zhang2011}.

EC is thus a popular family of potent algorithms that complements ML and DL to the benefit of all fields concerned. Further, there is a large and growing community of EC+ML practitioners.
We have used several EC open-source software packages over the years and have identified a large ``hole'' in the software landscape---there is a lacuna in the form of an EC package that is:

\begin{enumerate}
\item A comprehensive toolkit for running evolutionary algorithms.
\item Written in Python.
\item Can work with or without \tb{scikit-learn} (aka \tb{sklearn}), the most popular ML library for Python. To wit, the package should support both sklearn and standalone (non-sklearn) modes.
\item Designed with modern software engineering in mind.
\item Designed to support all popular EC paradigms: genetic algorithms (GAs), genetic programming (GP), evolution strategies (ES), coevolution, multi-objective, etc'.
\end{enumerate}

While there are several EC Python packages, none fulfill \textit{all} five requirements. Some are not written in Python, some are badly documented, some do not support multiple EC paradigms, and so forth. 
Importantly for the ML community, most tools do not intermesh with extant ML tools. Indeed, we have personally had experience with the hardships of combining EC tools with scikit-learn when doing evolutionary machine learning. We hope that by adhering to the above five pillars \tb{EC-KitY} will be deemed useful by a large community. 

\section{\tb{EC-KitY}}
\subsection{Setting up an evolutionary experiment}
\tb{EC-KitY} is available at \url{github.com/ec-kity/ec-kity/}, along with examples, tutorials, Google Colab notebooks, and more.
Installing it is simply done by:
\begin{lstlisting}[language=Python,upquote=true]
pip install eckity
\end{lstlisting}

As noted, \tb{EC-KitY} can work both in standalone, non-sklearn mode, and in sklearn mode.
In the following code examples, we use both modes for solving a symbolic regression problem, where the goal is to seek regressors of arbitrary complexity, i.e., beyond linear or polynomial ones \citep{Sipper2022esr}. 

We begin with the standalone mode, demonstrating how the user can run an EA with a mere three lines of code:
\begin{lstlisting}[language=Python,upquote=true]
from eckity.algorithms.simple_evolution import SimpleEvolution
from eckity.subpopulation import Subpopulation
from examples.treegp.non_sklearn_mode.symbolic_regression.sym_reg_evaluator import SymbolicRegressionEvaluator

algo = SimpleEvolution(Subpopulation(SymbolicRegressionEvaluator()))
algo.evolve()
print('algo.execute(x=2, y=3, z=4):', algo.execute(x=2, y=3, z=4))
\end{lstlisting}

An algorithm can operate on a single Subpopulation (as in the case of SimpleEvolution), two Subpopulations (e.g., in a coevolutionary setup), or $n$ ($>2$) Subpoplations (e.g., island model). Each Subpopulation may have its own evaluation function, which is why in the above example the SymbolicRegressionEvaluator is a parameter of Subpopulation. There are additional parameters of Subpopulation (e.g., genetic operators), and the algorithm (e.g., breeder and statistics). In the example, these parameters received the default values. The default function set is \texttt{[add, sub, mul, div]}, the default terminal set is \texttt{[\textquotesingle x\textquotesingle, \textquotesingle y\textquotesingle, \textquotesingle z\textquotesingle, 0, 1, -1]}, and there are additional parameters, as defined in the API~\cite{EC-KitY}. Below we show examples that demonstrate, among others, the use of non-default values. 

The above code takes 4 minutes to run in Google Colab; by adding a simple early-termination condition  (as in the last example below) runtime can be reduced to 9 seconds.

\newpage Here is an example of an evolved GP tree:
\begin{lstlisting}[language=Python,upquote=true]
add
   add
      z
      z
   add
      add
         z
         y
      add
         x
         y
\end{lstlisting}

The evolved tree's fitness (mean absolute error) is 9.41e-15. The code's output was:
\begin{lstlisting}[upquote=true]
algo.execute(x=2, y=3, z=4): 20
\end{lstlisting}

Running an EA in sklearn mode is just as simple (again, a symbolic-regression problem):
\begin{lstlisting}[language=Python,upquote=true]
from sklearn.datasets import make_regression
from sklearn.metrics import mean_absolute_error
from sklearn.model_selection import train_test_split

from eckity.algorithms.simple_evolution import SimpleEvolution
from eckity.creators.gp_creators.full import FullCreator
from eckity.genetic_encodings.gp.tree.utils import create_terminal_set
from eckity.sklearn_compatible.regression_evaluator import RegressionEvaluator
from eckity.sklearn_compatible.sk_regressor import SKRegressor
from eckity.subpopulation import Subpopulation

X, y = make_regression(n_samples=100, n_features=3)
terminal_set = create_terminal_set(X)
algo = SimpleEvolution(
         Subpopulation(creators=FullCreator(terminal_set=terminal_set),
                       evaluator=RegressionEvaluator()))
regressor = SKRegressor(algo)
X_train, X_test, y_train, y_test = train_test_split(X, y, test_size=0.2)
regressor.fit(X_train, y_train)
print('MAE on test set:', 
      mean_absolute_error(y_test, regressor.predict(X_test)))
\end{lstlisting}

In this example, we wrap the algorithm as a sklearn regressor and execute it from sklearn. The toolkit repository contains examples of more-advanced ways in which \tb{EC-KitY} interacts with sklearn, including grid search and pipeline. 

Running this code in Google Colab took 14 minutes without an early-termination condition, and the output was:
\begin{lstlisting}[upquote=true]
MAE on test set: 0.07
\end{lstlisting}

The user may wish to specify additional components of the EA rather than rely on default values; this can be readily achieved through the relevant constructor, e.g.: 

\begin{lstlisting}[language=Python,upquote=true]
algo = SimpleEvolution(
    Subpopulation(creators=
                    RampedHalfAndHalfCreator(init_depth=(2, 4),
                                             terminal_set=terminal_set,
                                             function_set=function_set,
                                             bloat_weight=0.0001),
                  population_size=200,
                  evaluator=SymbolicRegressionEvaluator(),
                  higher_is_better=False,
                  elitism_rate=0.05,
                  operators_sequence=[
                      SubtreeCrossover(probability=0.9, arity=2),
                      SubtreeMutation(probability=0.2, arity=1),
                      ERCMutation(probability=0.05, arity=1)
                  ],
                 ),
    breeder=SimpleBreeder(),
    max_workers=4,
    max_generation=500,
    statistics=BestAverageWorstStatistics(),
    termination_checker=
        ThresholdFromTargetTerminationChecker(optimal=0, threshold=0.001)
    )
\end{lstlisting}

Detailed API, tutorials, and several use cases are available at \cite{EC-KitY}.%changed to cite to make it fit into a single line. You can change it back if you prefer...

\subsection{Architecture}
As previously noted, we plan for \tb{EC-KitY} to support all main EC paradigms, including genetic algorithms, genetic programming, evolution strategies, coevolution, and evolutionary multi-objective optimization. To support this plethora of algorithms and genetic operators, we based our design on \tb{ECJ}, likely the most comprehensive EC package to date, which has been in use and under development for over two decades \citep{ECJ}. We did, however, veer away from \tb{ECJ} on a crucial point: Statistics can be drawn from any operation using event hooks, thus improving the decoupling and the separation of concerns. 

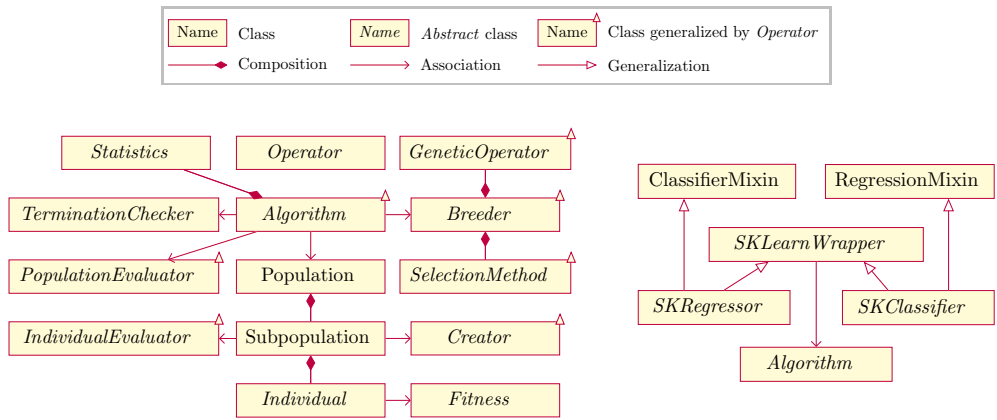
\begin{figure}
\centering

\def\y{0.5cm}
\def\sx{1.8cm}
\def\lx{2.2cm}
\def\lw{3.5cm}
\tikzset{
  umlcd style class/.append style={ % redefine this style
    text width=2.4cm,
    align=justify,
    % minimum width=2.6cm,
    % execute at begin node={\centring},
    execute at end node={\vphantom{g}},
  }
}
\tikzstyle{inherit}=[]
\tikzstyle{legend}=[anchor=north, umlcd style class, rectangle, text width=1cm, minimum height=0.1cm, minimum width=0.1cm]

% \begin{adjustbox}{cfbox=cyan 1pt,width=0.35\textwidth}
\begin{adjustbox}{cfbox=lightgray 1pt, width=0.65\textwidth}

\begin{tikzpicture}[,node distance=0.1cm]
\node[legend] (class-s) at (0,0) {Name};
\node[right=of class-s] (class-t) {Class};

\node[legend, right=1.5cm of class-t] (abstract-s) {\emph{Name}};
\node[right=of abstract-s] (abstract-t) {\emph{Abstract} class};

\node[legend, inherit, right=0.3cm of abstract-t] (operator-s) {Name};
\node[right=of operator-s] (operator-t) {Class generalized by \emph{Operator}};
\node[umlcd style, isosceles triangle, isosceles triangle apex angle=45,draw,rotate=90,minimum size=0.15cm,inner sep=0mm] at (operator-s.north east){};

\draw[umlcd style, -open triangle 45]([yshift=-0.7cm]operator-s.west) -- ([yshift=-0.7cm]operator-s.east) node (extends-s) {};
\node[right=of extends-s.center](extends-t){Generalization};

\draw[umlcd style, -diamond]([yshift=-0.7cm]class-s.west) -- ([yshift=-0.7cm]class-s.east) node (composition-s) {};
\node[right=of composition-s.center](composition-t){Composition};

\draw[umlcd style, ->]([yshift=-0.7cm]abstract-s.west) -- ([yshift=-0.7cm]abstract-s.east) node (use-s) {};
\node[right=of use-s.center](use-t) {Association};

% \draw[dashed] ([shift=({-0.3,-0.2})]current bounding box.south west) -- ([shift=({0.3,0})]current bounding box.south east);
\end{tikzpicture}
\end{adjustbox}
\vspace{3ex}

\begin{subfigure}[c]{.55\textwidth}
\begin{adjustbox}{width=\textwidth}
\begin{tikzpicture}
\begin{abstractclass}[]{Operator}{0,0}\end{abstractclass}
% \pgfpointdiff{\pgfpointanchor{Operator}{north}}{\pgfpointanchor{Operator}{south}}
% \pgfmathsetmacro\abstractheight{\csname pgf@y\endcsname}

\begin{abstractclass}[yshift=-\y,inherit]{Algorithm}{Operator.south}\inherit{Operator}\end{abstractclass}
\begin{abstractclass}[xshift=\sx,inherit]{Breeder}{Algorithm.north east}\inherit{Operator}\end{abstractclass}
\begin{class}[yshift=-\y]{Population}{Algorithm.south}\end{class}
\begin{abstractclass}[text width=\lw, xshift=-\lx,inherit]{PopulationEvaluator}{Population.north west}\inherit{Operator}\end{abstractclass}
\begin{class}[yshift=-\y]{Subpopulation}{Population.south}\end{class}
\begin{abstractclass}[yshift=-\y]{Individual}{Subpopulation.south}\end{abstractclass}
\begin{abstractclass}[xshift=\sx]{Fitness}{Individual.north east}\end{abstractclass}
\begin{abstractclass}[text width=\lw, xshift=-\lx,inherit]{IndividualEvaluator}{Subpopulation.north west}\inherit{Operator}\end{abstractclass}
\begin{abstractclass}[text width=\lw, xshift=-\lx]{TerminationChecker}{Algorithm.north west}\end{abstractclass}
\begin{abstractclass}[xshift=\sx,inherit]{Creator}{Subpopulation.north east}\inherit{Operator}\end{abstractclass}
\begin{abstractclass}[text width=2.8cm, xshift=\sx,inherit]{SelectionMethod}{Population.north east}\inherit{Operator}\end{abstractclass}
\begin{abstractclass}[text width=2.8cm, xshift=\sx,inherit]{GeneticOperator}{Operator.north east}\inherit{Operator}\end{abstractclass}
\begin{abstractclass}[xshift=-\sx]{Statistics}{Operator.north west}\end{abstractclass}

\unidirectionalAssociation{Algorithm}{}{}{Population}{}{}
\unidirectionalAssociation{Algorithm}{}{}{TerminationChecker}{}{}
\unidirectionalAssociation{Algorithm}{}{}{Breeder}{}{}
\unidirectionalAssociation{Individual}{}{}{Fitness}{}{}
\unidirectionalAssociation{Algorithm}{}{}{PopulationEvaluator}{}{}
\unidirectionalAssociation{Subpopulation}{}{}{IndividualEvaluator}{}{}
\unidirectionalAssociation{Subpopulation}{}{}{Creator}{}{}
\composition{Algorithm}{}{}{Statistics}{}
\composition{Population}{}{}{Subpopulation}
\composition{Breeder}{}{}{SelectionMethod}
\composition{Breeder}{}{}{GeneticOperator}
\composition{Subpopulation}{}{}{Individual}
\end{tikzpicture}
\end{adjustbox}
% \caption{}
% \label{fig:sub-first}
\end{subfigure}\qquad
\begin{subfigure}[c]{.346\textwidth}
\begin{adjustbox}{width=\textwidth}
% include second image
\begin{tikzpicture}
\begin{abstractclass}[text width=\lw]{SKLearnWrapper}{0,0}\end{abstractclass}
\begin{abstractclass}[yshift=-{3*\y}]{Algorithm}{SKLearnWrapper.south}\end{abstractclass}
\begin{abstractclass}[xshift=\sx, yshift=-\y]{SKClassifier}{SKLearnWrapper.south}\end{abstractclass}
\begin{abstractclass}[xshift=-\sx, yshift=-\y]{SKRegressor}{SKLearnWrapper.south}\end{abstractclass}
\begin{class}[text width=2.5cm, yshift=\y, anchor=south west]{ClassifierMixin}{SKLearnWrapper.north -| SKRegressor.west}\end{class}
\begin{class}[text width=2.7cm, yshift=\y, anchor=south east]{RegressionMixin}{SKLearnWrapper.north -| SKClassifier.east}\end{class}

\draw[umlcd style, ->](SKLearnWrapper) -- (Algorithm) node[left,pos=0.9]{};%{wraps};
\draw [umlcd style, -open triangle 45] (SKRegressor.150) -- (SKRegressor.150 |- ClassifierMixin.south);
\draw [umlcd style, -open triangle 45] (SKClassifier.30) -- (SKClassifier.30 |- RegressionMixin.south);

\draw [umlcd style, -open triangle 45] (SKRegressor.60) -- (SKLearnWrapper.200);
\draw [umlcd style, -open triangle 45] (SKClassifier.150) -- (SKLearnWrapper.340);
\end{tikzpicture}
\end{adjustbox}
% \caption{}
% \label{fig:sub-second}
\end{subfigure}

\caption{The general architecture of \tb{EC-KitY} (left), and the added components for sklearn mode (right).
}
\label{fig:architecture}
\end{figure}

The architecture of \tb{EC-KitY} is depicted in Figure \ref{fig:architecture}. The \texttt{Algorithm} class serves as the entry point to \tb{EC-KitY}. It contains all the needed information for an experiment and acts on the population using the \texttt{Breeder} class. Several classes extend the \texttt{Operator} class (these are denoted by a small triangle); they emit events that can be intercepted by \texttt{Statistics}.
The same \tb{EC-KitY} code can be used both in standalone mode (Figure~\ref{fig:architecture}, left) and in sklearn mode (Figure~\ref{fig:architecture}, right). The latter is achieved by wrapping an \texttt{Algorithm} instance with an \texttt{SKLearnWrapper}.

\tb{EC-KitY} is designed to be easily extended: it is heavily over-architected, with many hooks that facilitate system modification and enhancement. 
The architecture supports the following features: execution in the cloud or in a cluster, multiple statistics, multithreaded evaluation, replicability standards (all of which have already been implemented), as well as checkpointing and logging facilities (under active development).
%The architecture supports job handling, checkpointing, multiple statistics, and applicability standards (already implemented), as well as multithreaded evaluation and breeding, and logging facilities (under active development) 

\section{Comparison with Other Packages}
Through considerable hands-on experience with EC open-source software over many years, and through additional extensive research, we have identified the following eight tools, whose major features are compared in Table~\ref{tab:compare}, along with \tb{EC-KitY} \citep{EC-KitY}:
\tb{gplearn} \citep{gplearn}, 
\tb{geatpy} \citep{geatpy}, 
\tb{DEAP} \citep{DEAP}, 
\tb{Platypus} \citep{platypus}, 
\tb{ECJ} \citep{ECJ}, 
\tb{Jenetics} \citep{Jenetics}, 
\tb{KEEL} \citep{KEEL}, 
\tb{HeuristicLab} \citep{HeuristicLab}.

Only two packages are written in Python and are also sklearn-compatible: \tb{geatpy} and \tb{gplearn}. However, as shown in Table~\ref{tab:compare}, they lack many important features that \tb{EC-KitY} possesses. 

\begin{table}
\caption{Feature comparison of \tb{EC-KitY} with extant software packages.
        \ck{\bf :} feature exists,
        \ckp{\bf :} feature planned soon,
        \ckm{\bf :} feature mostly lacking,
        \xm{\bf :} feature completely lacking.}
\label{tab:compare}
\centering
\resizebox{0.99\textwidth}{!}{
    \begin{tabular}{|r|c|c|c|c|c|c|c|c|c|}
    \hline
        {\bf Feature}               & \tb{EC-KitY}  & \tb{geatpy}   & \tb{gplearn}  & \tb{DEAP}  & \tb{Platypus}  & \tb{ECJ}  & \tb{Jenetics}  & \tb{ KEEL} & \tb{HeuristicLab} \\ \hline
        Language                    & \python       & \python       & \python       & \python    & \python        & \java     & \java          & \java      & \csharp           \\ \hline
        sklearn-compatible          & \ck           & \ck           & \ck           & \xm        & \xm            & \xm       & \xm            & \xm        & \xm               \\ \hline
        SE Design                   & \ck           & \ckm          & \xm           & \xm        & \xm            & \ck       & \ck            & \ckm       & \ck               \\ \hline
        GA representations          & \ck           & \ck           & \xm           & \ck        & \ck            & \ck       & \ck            & \ck        & \ck               \\ \hline
        GP representations          & \ck           & \xm           & \ck           & \ck        & \xm            & \ck       & \ckm           & \ck        & \ck               \\ \hline
        User-defined operators      & \ck           & \xm           & \xm           & \ck        & \ck            & \ck       & \ck            & \xm        & \ck               \\ \hline
        User-defined representation & \ck           & \xm           & \xm           & \ck        & \ck            & \ck       & \ck            & \xm        & \ck               \\ \hline
        Coevolution                 & \ckp          & \xm           & \xm           & \ck        & \xm            & \ck       & \xm            & \ck        & \xm               \\ \hline
        Multiobjective              & \ck          & \ck           & \xm           & \ck        & \ck            & \ck       & \ck            & \ck        & \ck               \\ \hline
        Statistics                  & \ck           & \ck           & \ck           & \ck        & \ckm           & \ck       & \ck            & \ck        & \ck               \\ \hline
        Documentation               & \ck           & \ckm          & \ck           & \ckm       & \ckm           & \ck       & \ck            & \ckm       & \ck               \\ \hline
        API                         & \ck           & \ckm          & \ck           & \ck        & \xm            & \ck       & \ck            & \xm        & \ck               \\ \hline
        Latest version              & \bl{2022}     & \bl{2022}     & \rd{2019}     & \rd{2019}  & \rd{2020}      & \rd{2019} & \bl{ 2022}     & \rd{2015}  & \bl{2022}         \\ \hline
    \end{tabular}
}
\end{table}

\section{Concluding Remarks}
The \tb{EC-KitY} library is under active development. Presently, we plan to add a variety of evolutionary algorithms, individual types, genetic operators, and tests.
We are also working on extending and enhancing the documentation. We wish to broaden both the user and developer communities by implementing more sample use cases from different domains and by encouraging developers to contribute. 

We recently taught a course in which 48 students worked in groups of two or three, submitting a total of 22 projects that used \tb{EC-KitY} to solve a diverse array of complex problems, including evolving Flappy Bird agents, evolving blackjack strategies, evolving Super Mario agents, evolving chess players, and solving problems such as maximum clique and vehicle routing. 
\tb{EC-KitY} proved quite up to the tasks.

Given the popularity and diverse applicability of evolutionary algorithms, we hope that \tb{EC-KitY} finds multitudinous and beneficial uses.

\section*{Acknowledgements}
This research was partially supported by the following grants: 2714/19 from the Israeli Science Foundation; Israeli Smart Transportation Research Center (ISTRC); Israeli Council for Higher Education (CHE) via the Data Science Research Center, Ben-Gurion University of the Negev, Israel.

{\footnotesize
\setlength{\bibsep}{3pt}
\bibliographystyle{elsarticle-num-names} % elsarticle-harv, elsarticle-num-names, elsarticle-num
\bibliography{refs}
}

% \section*{Illustrative Examples}
% Optional : you may include one explanatory  video that will appear next to your article, in the right hand side panel. (Please upload any video as a single supplementary file with your article. Only one MP4 formatted, with 50MB maximum size, video is possible per article. Recommended video dimensions are 640 x 480 at a maximum of 30 frames / second. Prior to submission please test and validate your .mp4 file at  \url{http://elsevier-apps.sciverse.com/GadgetVideoPodcastPlayerWeb/verification} . This tool will display your video exactly in the same way as it will appear on ScienceDirect. )

\section*{Required Metadata}
\label{sec:metadata}

\section*{Current code version}
\label{}
% Ancillary data table required for subversion of the codebase. Kindly replace examples in right column with the correct information about your current code, and leave the left column as it is.

\begin{table}[!h]
\begin{tabular}{|l|p{6.5cm}|p{6.5cm}|}
\hline
% \textbf{Nr.} & \textbf{Code metadata description} & \textbf{Please fill in this column} \\
% \hline
C1 & Current code version & v0.3.0 \\
\hline
C2 & Permanent link to code/repository used for this code version & \url{https://github.com/EC-KitY/EC-KitY} \\
\hline
C3  & Permanent link to Reproducible Capsule & \url{https://codeocean.com/capsule/8740191/tree} \\
\hline
C4 & Legal Code License   & BSD 3-Clause License \\
\hline
C5 & Code versioning system used & git \\
\hline
C6 & Software code languages, tools, and services used & Python \\
\hline
C7 & Compilation requirements, operating environments \& dependencies & 
numpy ($>=1.14.6$),
pandas\newline ($>=0.25.0$),
overrides ($>= 6.1.0$). For sklearn mode: scikit-learn\newline ($>=0.24.2$) \\
\hline
C8 & Link to developer documentation/manual & \textit{API}---\url{https://api.eckity.org/eckity.html}\newline
\textit{Tutorials}---\url{https://github.com/EC-KitY/EC-KitY/wiki/Tutorials}\newline
\textit{Examples}---\url{https://github.com/EC-KitY/EC-KitY/tree/main/examples} \\
\hline
C9 & Support email for questions &  \href{mailto:sipper@bgu.ac.il}{sipper@bgu.ac.il}\\
\hline
\end{tabular}
\caption{Code metadata}
% \label{} 
\end{table}

\section*{Current executable software version}
\label{}

\begin{table}[!h]
\begin{tabular}{|l|p{6.5cm}|p{6.5cm}|}
% \hline
% \textbf{Nr.} & \textbf{(Executable) software metadata description} & \textbf{Please fill in this column} \\
\hline
S1 & Current software version & v0.3.0 \\
\hline
S2 & Permanent link to executables of this version  & \url{https://pypi.org/project/eckity} \\
\hline
S3  & Permanent link to Reproducible Capsule & \url{https://codeocean.com/capsule/8740191/tree}\\
\hline
S4 & Legal Software License & BSD 3-Clause License \\
\hline
S5 & Computing platforms/Operating Systems & Cross-platform \\
\hline
S6 & Installation requirements \& dependencies & numpy ($>=1.14.6$),
pandas\newline ($>=0.25.0$),
overrides ($>= 6.1.0$). For sklearn mode: scikit-learn\newline ($>=0.24.2$)\\
\hline
S7 & Link to user manual & \textit{API}---\url{https://api.eckity.org/eckity.html}\newline
\textit{Tutorials}---\url{https://github.com/EC-KitY/EC-KitY/wiki/Tutorials}\newline
\textit{Examples}---\url{https://github.com/EC-KitY/EC-KitY/tree/main/examples}  \\
\hline
S8 & Support email for questions & \href{mailto:sipper@bgu.ac.il}{sipper@bgu.ac.il}\\
\hline
\end{tabular}
\caption{Software metadata}
% \label{} 
\end{table}

\end{document}